%% file: main.tex
\documentclass[letterpaper, 10 pt, conference]{ieeeconf}  

\IEEEoverridecommandlockouts                              

\usepackage{cite}
\usepackage{amsmath,amssymb,amsfonts}
\usepackage{algorithmic}
\usepackage{graphicx}
\usepackage{textcomp}
\usepackage{xcolor}
\usepackage{algorithm}
\usepackage{array}
\usepackage[caption=false,font=normalsize,labelfont=sf,textfont=sf]{subfig}
\usepackage{textcomp}
\usepackage{utfsym}
\usepackage{pifont}
\usepackage{stfloats}
\usepackage{url}
\usepackage{verbatim}
\usepackage{hyperref}

\hypersetup{
	colorlinks=true,
	linkcolor=[rgb]{1,0.058,0.121},
	filecolor=cyan,  
	urlcolor=[rgb]{0.948,0.008,0.56},
	citecolor=[rgb]{0.298,1,0.309},
}

\usepackage{multirow}
\usepackage{booktabs}
\usepackage{makecell}
\usepackage{bm}

\definecolor{LightCyan}{rgb}{0.88,1,1}

\definecolor{sgreen}{RGB}{30, 150, 30}

\usepackage{bm}
\usepackage{mathrsfs}
\usepackage{booktabs}
\usepackage{multicol}
\usepackage{color}
\usepackage{hyperref}
\usepackage{float}
\usepackage{subfig}
\usepackage[table]{xcolor}
\usepackage{graphicx}
\usepackage{multirow}
\let\labelindent\relax
\usepackage{enumitem}
\usepackage{bbding}
\usepackage{mathrsfs}
\usepackage{color}
\usepackage{hyperref}
\usepackage{float}
\usepackage{colortbl}
\usepackage{rotating}
\usepackage{array}
\definecolor{LightCyan}{rgb}{0.88,1,1}
\definecolor{cvprblue}{rgb}{0.21,0.49,0.74}

\usepackage{xcolor}
\usepackage{adjustbox}
\usepackage{tabularx}
\usepackage{etoolbox}
\usepackage{makecell}
\usepackage{pifont}
\usepackage{wrapfig}
\usepackage{cutwin}
\usepackage{alltt}

\title{\textbf{AeroRAG: Structured Multimodal Retrieval-Augmented LLM \\ for Fine-Grained Aerial Visual Reasoning}
}

\author{\hspace{1.1em}Junxiao Xue$^{1\dagger}$, Quan Deng$^{2,1\dagger}$, Tingqi Hu$^{3}$, Meicong Si$^{3}$, Xinyi Yin$^{3}$, Yunyun Shi$^{4}$, and Xuecheng Wu$^{4\ast\ddagger}$
\thanks{*This work was supported by the Central Government Guiding Local Science and Technology Development Fund under Grant No. 2026ZY04001.}
\thanks{$^{1}$Research Center for Space Computing System, Zhejiang Lab, Hangzhou 311100, China (E-mail: {\small xuejx@zhejianglab.cn});}
\thanks{$^{2}$Hangzhou Institute for Advanced Study, University of Chinese Academy of Sciences, Hangzhou 311000, China (E-mail: {\small dengquan23@mails.ucas.ac.cn});}
\thanks{$^{3}$School of Cyber Science and Engineering, Zhengzhou University, Zhengzhou 450002, China (E-mail: {\small \{htqiserendipity, smc\_ggjxb, yinxinyi\}@stu.zzu.edu.cn}).}
\thanks{$^{4}$School of Computer Science and Technology, Xi'an Jiaotong University, Xi'an 710049, China (E-mail: {\small yunyunshi@stu.xjtu.edu.cn; wuxc3@ieee.org});}
\thanks{Work done during Quan Deng's research internship at Zhejiang Lab.}
\thanks{$^{\ast}$Corresponding author: Xuecheng Wu.}
\thanks{$^{\dagger}$Equal contributions. $^{\ddagger}$Project lead.}
}

\begin{document}
\maketitle

\thispagestyle{empty}
\pagestyle{empty}

\input{Sec/0_abstract}

\begin{figure}[t!]
    \centering
    \includegraphics[width=\linewidth]{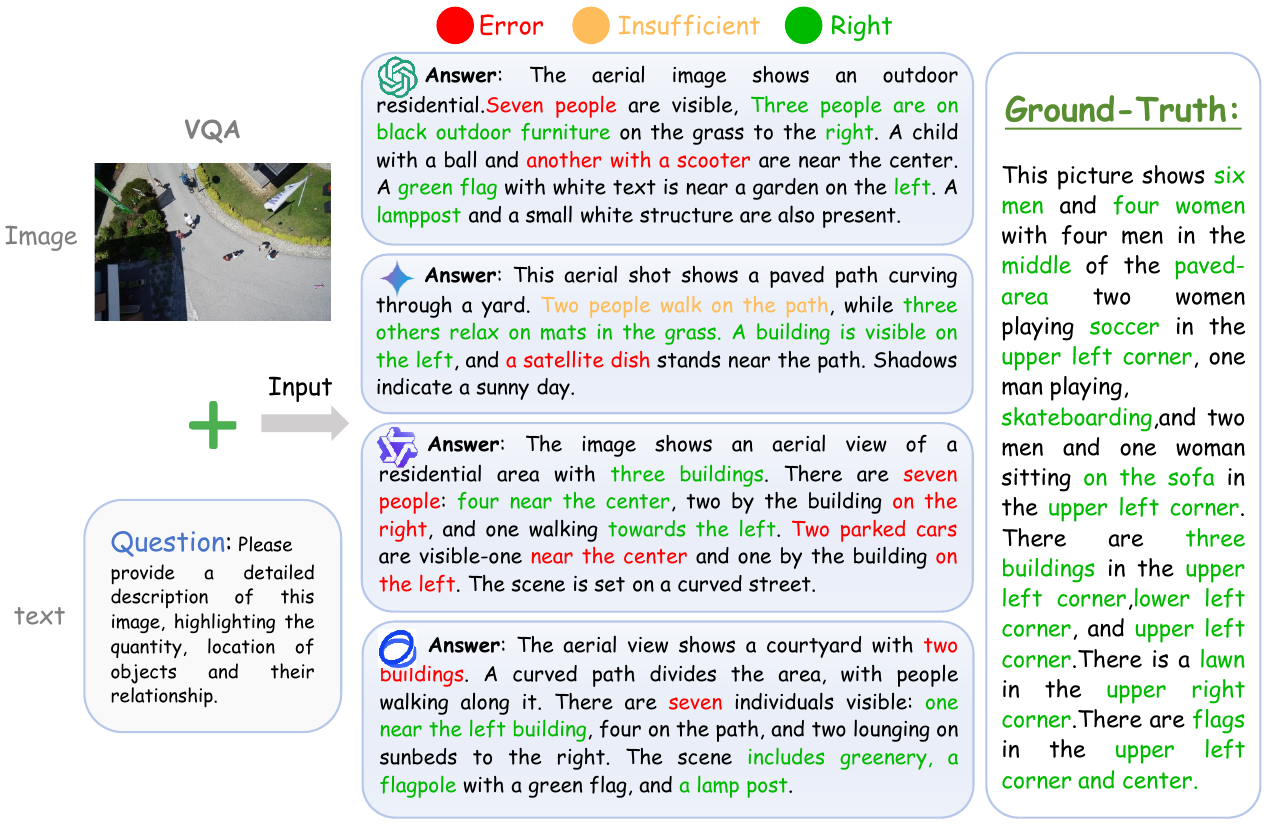}
    \caption{The qualitative comparison on challenging aerial VQA examples. Existing MLLMs often miss small objects or confuse counts and spatial relations, whereas our proposed multimodal RAG framework provides more accurate and visually grounded answers.}
    \label{fig:SamplePictures}
\end{figure}

\input{Sec/1_introduction}
\input{Sec/2_relate_work}

\input{Sec/3_methodology}
\input{Sec/4_experiments}
\input{Sec/5_conclusion}

\bibliographystyle{IEEEbib}
\bibliography{smc2026references}

\end{document}

%% file: Sec/0_abstract.tex
\begin{abstract}
Recent progress in multimodal large language models (MLLMs) has yet to resolve the challenge of reliable visual question answering in aerial imagery. In such scenes, task-critical evidence is often carried by small objects, explicit quantities, coarse locations, and inter-object relations, whereas conventional dense visual-token representations are not well aligned with these structured semantics. To address this interface mismatch, we propose AeroRAG, a scene-graph-guided multimodal retrieval-augmented generation framework for visual question answering. The framework first converts an input image into structured visual knowledge, including object categories, quantities, spatial locations, and semantic relations, and then retrieves query-relevant semantic chunks to construct compact prompts for a text-based large language model. Rather than relying on direct reasoning over dense visual tokens, our method introduces a more explicit intermediate interface between perception and language reasoning. Experiments on the AUG aerial dataset and the general-domain VG-150 benchmark show consistent improvements over six strong MLLM baselines, with the largest gains observed in dense aerial scenes and relation-sensitive reasoning. We further evaluate the framework on VQAv2 to demonstrate that the proposed interface remains compatible with standard visual reasoning settings. These results suggest that structured retrieval is a practical design direction for deployment-oriented and grounded visual reasoning systems.
\end{abstract}

%% file: Sec/1_introduction.tex
\section{Introduction}

Multimodal large language models (MLLMs) have become essential frameworks for merging visual and linguistic understanding, fueled by various advanced models~\cite{xue2026towards,wu2025vic,wu2025avf}. These approaches connect pre-trained vision encoders with large language models (LLMs), achieving outstanding performance in general visual question answering (VQA)~\cite{shao2023prompting,zhang2025hkd4vlm} and image captioning~\cite{NEURIPS2023_804b5e30}. At the same time, there is increasing interest in deploying such models in mission critical visual analysis scenarios, especially in aerospace and aerial imagery applications such as remote sensing, environmental monitoring, urban planning, and autonomous inspection. In these settings, the requirement is not only strong perception, but also reliable and grounded reasoning over complex scenes.

However, directly applying existing MLLMs to aerial VQA remains challenging. As illustrated in Fig.~\ref{fig:SamplePictures}, even advanced models such as GPT-4o~\cite{gpt4} may miscount overlapping objects and fail on relation sensitive questions when scenes contain small, densely distributed, or partially occluded entities. In aerial imagery, such errors are not merely minor deviations in language generation. They can directly affect object counting, coarse localization, and scene interpretation, thereby reducing the reliability of downstream decision making. This observation suggests that the main bottleneck is not simply whether a model can see the scene, but whether task critical visual evidence can be organized and exposed to the reasoning module in a suitable form.

We argue that this difficulty can be understood as an interface mismatch between standard MLLMs and the requirements of aerial VQA. Conventional MLLMs typically rely on dense and largely unstructured visual token sequences, whereas aerial reasoning often depends on sparse but decision critical semantics, including object categories, explicit quantities, coarse spatial locations, and inter-object relations. In dense aerial scenes, these structured cues are difficult to access reliably from raw visual tokens alone. As a result, the reasoning module must infer fine-grained scene semantics from a representation that is not naturally aligned with the structure of the task. From a systems perspective, the challenge is therefore not only model capacity, but also how visual evidence is represented and delivered across the perception-to-reasoning pipeline.

To address this issue, we propose AeroRAG, a pluggable multimodal retrieval-augmented generation framework for aerial visual question answering. Instead of asking the language model to reason directly over dense visual tokens, AeroRAG first converts the input image into scene-graph-based structured knowledge that explicitly captures object categories, quantities, coarse locations, and inter-object relations. It then performs query-conditioned retrieval over this structured representation and assembles the most relevant semantic chunks into compact textual prompts for a text-based LLMs. In this way, the proposed framework introduces a more explicit intermediate interface between perception and language reasoning. In our implementation, this design provides grounded and inspectable evidence for downstream answering while avoiding additional domain-specific multimodal pre-training.

Recent advances in multimodal RAG, such as fine-grained retrieval for VQA \cite{zhang2025fine}, style aware image captioning, and OCR or detection based retrieval pipelines \cite{bag2024rag, kumar2024overcoming}, have demonstrated the potential of augmenting LLMs with external visual evidence. However, these existing multimodal RAG methods primarily rely on converting images into shallow textual representations and appending them to dense visual inputs. This shallow extraction is fundamentally insufficient for dense aerial scenes, where task critical cues rely on the joint reasoning of object categories, exact quantities, spatial locations, and complex inter object relations. To address this interface mismatch, AeroRAG introduces a paradigm shift: rather than appending fragmented text to dense visual tokens, we propose utilizing a scene graph as the core intermediate interface. By explicitly transforming unstructured pixels into structured semantic chunks, encapsulating relational triples and spatial layouts, our framework provides a more grounded, inspectable, and noise-filtered context for the language model, fundamentally distinguishing AeroRAG from conventional multimodal RAG pipelines.

In conclusion, the main contributions of this paper are summarized as follows:
\begin{itemize}
    \item We reformulate aerial VQA as a structured visual grounding and retrieval problem, highlighting the importance of object categories, quantities, coarse locations, and inter-object relations in dense aerial scenes.
    
    \item We develop AeroRAG, a pluggable framework that converts images into scene-graph-based structured knowledge and uses query-conditioned retrieval to construct compact prompts for a text based LLMs. In our implementation, this design offers a deployment-oriented alternative to domain-specific multimodal retraining.
    
    \item Experiments on AUG and VG-150 show consistent improvements over six strong MLLMs baselines, with particularly clear gains in dense aerial scenes. Additional results on VQAv2 further suggest that the proposed interface remains applicable beyond the aerial setting.
\end{itemize}

%% file: Sec/2_relate_work.tex
\section{Related work}

\subsection{Multimodal Large Language Models}

MLLMs have become a leading framework for unifying vision and language, achieved by extending LLMs to handle both visual and textual inputs~\cite{multimodal, llm_survey}. Representative frameworks typically combine a pre-trained LLMs backbone, such as LLaMA~\cite{llama}, Qwen~\cite{qwen}, and Vicuna~\cite{vicuna}, with visual encoders such as CLIP~\cite{clip}, BLIP~\cite{blip}, ViT~\cite{vit}, and Q-Former~\cite{blip2}, often through lightweight adapter modules~\cite{adapter}. These systems have demonstrated strong performance in general visual-language tasks, including image captioning and VQA.

However, most existing MLLMs still represent an image as a dense sequence of visual tokens and rely on the language model to infer task-relevant evidence from the full multimodal context. This strategy is effective for coarse scene understanding, but it becomes less suitable when the target semantics are sparse yet decision critical, such as exact object counts, coarse locations, and inter object relations. The limitation is particularly evident in aerial imagery, where objects are often tiny, densely distributed, and partially occluded. In such settings, dense visual token representations may make it difficult for the reasoning module to reliably access fine-grained structured evidence, thereby increasing the risk of hallucinated or spatially inconsistent answers.

Currently, parameter-efficient fine-tuning (PEFT) methods~\cite{peft}, such as LoRA~\cite{lora}, Prefix Tuning~\cite{prefix}, and other PEFT strategies, can impressively reduce adaptation cost, but they do not fundamentally redesign the visual-language interface. As a result, they do not explicitly address the problem of structured grounding in dense aerial reasoning. This motivates us to explore an alternative interface, where the image is first converted into structured visual knowledge before being consumed by the language model.

\subsection{Retrieval-Augmented Generation}

By fetching external knowledge at inference time, retrieval-augmented generation (RAG) enhances the quality of the generated output~\cite{fan2024survey,jeong2023generative}. In multimodal settings, this idea can be extended by transforming visual observations into indexable knowledge units, allowing the language model to access explicit grounded context instead of relying solely on implicit multimodal representations. Prior studies have explored multimodal RAG in tasks such as visual question answering~\cite{zhang2025fine}, image captioning~\cite{bazi2025ragcap, sarto2024towards}, and cross-modal retrieval~\cite{zheng2025retrieval}. External perception tools, such as OCR~\cite{bag2024rag} and object detection models~\cite{kumar2024overcoming}, further support this direction by converting visual content into retrievable textual evidence.

Nevertheless, directly applying existing multimodal RAG pipelines to dense aerial scenes remains insufficient. First, many methods focus on relatively shallow visual extraction, such as isolated objects or text regions, without using scene level structured relations as the principal interface between perception and reasoning. This is a critical limitation for aerial VQA, where category, quantity, location, and relation often need to be considered jointly. Second, in many pipelines, retrieved information is appended to already dense multimodal inputs rather than replacing them as the main reasoning interface. As a result, retrieval may provide additional evidence, but it does not fully resolve the difficulty of organizing fine-grained scene semantics into a compact and directly usable form for language reasoning.

Our work is motivated by this gap and fundamentally differs from existing multimodal RAG methods in two key aspects. First, unlike traditional pipelines that rely on shallow visual extraction, such as extracting isolated object regions or OCR text, AeroRAG employs a scene-graph-based structural representation as the primary reasoning interface. We explicitly model and preserve inter-object relational triples $(o_i, p, o_j)$, coarse locations, and quantities, which are critical for dense aerial reasoning but often lost in prior methods that only process independent visual patches. Second, while many existing multimodal RAG approaches append retrieved textual evidence to already dense visual tokens, our framework completely replaces the dense visual token input with compact, query-relevant structured semantic chunks. By doing so, AeroRAG not only avoids the visual token explosion in dense scenes but also enforces the language model to reason strictly over explicitly grounded visual relationships, thereby significantly mitigating hallucination.

%% file: Sec/3_methodology.tex
\section{Methodology}

\begin{figure*}[t!]
    \centering
    \includegraphics[width=\linewidth]{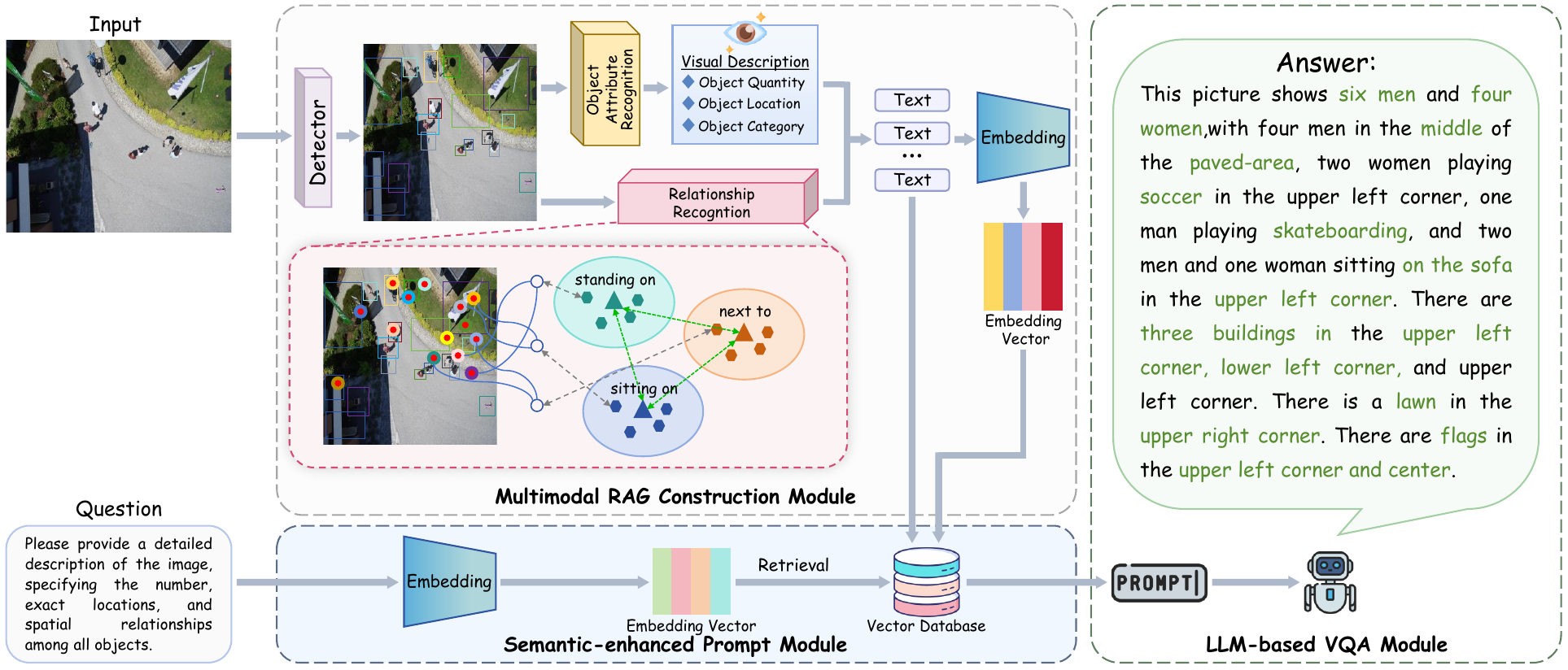}
    \caption{Overview of the enhanced multimodal RAG-LLM framework for visual question answering, comprising three core components: (1) Multimodal RAG Construction enabling structured visual encoding through knowledge graph generation, (2) Semantic-Enhanced Prompt Generation for cross-modal context alignment, and (3) LLM-Based VQA Engine integrating domain knowledge for response synthesis with visual grounding.}
    \label{fig:framework}
\end{figure*}

As depicted in Fig.~\ref{fig:framework}, our structured multimodal RAG framework addresses fine-grained visual reasoning through three modules: (1) \emph{Multimodal RAG Construction} for structured visual encoding, (2) \emph{Semantic-Enhanced Prompt Generation} for context alignment, and (3) \emph{LLM-Based VQA} for knowledge-guided response synthesis. 

\subsection{Multimodal RAG Construction Module}
The multimodal RAG construction module establishes structured visual-semantic representations through three synergistic components: 
(i) hierarchical visual attribute extraction, 
(ii) context-aware relationship modeling, and 
(iii) semantic-constrained vector indexing. 
This pipeline effectively transforms raw visual inputs into query ready multimodal knowledge representations.

\noindent\textbf{Structured Visual Understanding:}
For a given image $I$, a scene graph $G_I$ is first generated by the framework, encapsulating the objects within the image, their intrinsic attributes, and the spatial relationships that hold between them. The formal definition of the scene graph is:
\begin{equation}
G_I = (O, \mathcal{P}, \mathcal{R}),
\end{equation}
where $O=\{o_1,o_2,\dots,o_N\}$ represents the set of \textit{objects} (nodes), $\mathcal{P}$ is the set of \textit{predicates}, and $\mathcal{R} \subseteq O \times \mathcal{P} \times O$ is the set of semantic triples that capture relationships between pairs of objects.

To construct $G_I$, we initially employ Faster R-CNN~\cite{girshick2015fast} to extract object proposals and corresponding feature vectors. For each detected object $o_i \in O$, we assign specific attributes to create a comprehensive structured representation:
\begin{itemize}
    \item \textbf{Category and Quantity:} Each object is assigned a semantic category label $l_i \in \mathcal{L}$. Furthermore, we aggregate the total number of instances $N_{l_i}$ for each category to provide a global quantitative overview of the scene.
    \item \textbf{Bounding Box:} The spatial dimensions are defined by $b_i = (x_{\text{min}}, y_{\text{min}}, x_{\text{max}}, y_{\text{max}})$, locating the exact coordinates of $o_i$ within $I$.
    \item \textbf{Spatial Location:} We compute the center point $c_i = (\frac{x_{\text{min}} + x_{\text{max}}}{2}, \frac{y_{\text{min}} + y_{\text{max}}}{2})$ of $b_i$. The image $I$ is then uniformly divided into a $3 \times 3$ grid. The center point $c_i$ is mapped to one of these nine regions to concisely describe its relative spatial location.
\end{itemize}

The structured representation facilitates a deeper contextual understanding and serves as a robust foundation for subsequent reasoning tasks.

\subsubsection{Relationship Identification} 
To explicitly construct the semantic triples for relationship reasoning, we employ the Prototype-based Embedding Network (PENET)~\cite{penet}. Unlike direct feature concatenation, PENET first generates object representations for the subject and object, and then predicts the predicate based on their pairwise interaction. Each relationship is formalized as a semantic triple $(o_i, p, o_j) \in R$, where $o_i$, $o_j$, and $p$ represent the subject, object, and predicate, respectively.

\textbf{Object Representation:} 
For each detected object, whether it serves as the subject $o_i$ or the object $o_j$, its representation is decomposed into a base semantic prototype representing the general class concept and an instance-specific visual deviation capturing the unique visual characteristics in the current image. Their formulations are:
\begin{equation}
    o_i = W_i t_i + v_i,
\end{equation}
\begin{equation}
    o_j = W_j t_j + v_j,
\end{equation}
\begin{equation}
    v_i = \sigma \left( \text{FC}(W_i t_i, M(e_i)) \right) \odot M(e_i),
\end{equation}
where $\sigma(\cdot)$ is the ReLU activation function, $\text{FC}(\cdot)$ is a fully connected layer, and $M(\cdot)$ is a visual-to-semantic projection function. The visual features $e_i$ are extracted from the Faster R-CNN detection module. The vector $v_j$ is computed symmetrically to $v_i$.

\textbf{Predicate Inference and Feature Fusion:} 
Once the representations of the subject $o_i$ and object $o_j$ are determined, we proceed to model the predicate $p$. The predicate representation also follows a prototype-deviation decomposition, but crucially, its visual deviation $u_p$ depends on the interaction between the subject and the object:
\begin{equation}
    p = W_p t_p + u_p,
\end{equation}
\begin{equation}
    u_p = \sigma \left( \text{FC}(G(o_i, o_j), M(e_p)) \right) \odot M(e_p),
\end{equation}
where $t_p$ is the semantic prototype of the predicate, and $e_p$ is the visual feature of its region. To model spatial and semantic interactions between subject and object, we design a feature fusion function $G(o_i, o_j)$:
\begin{equation}
    G(o_i, o_j) = \text{ReLU}(o_i + o_j) - (o_i - o_j)^2.
\end{equation}
This fusion mechanism is designed based on two motivations: (1) the addition term $\text{ReLU}(o_i + o_j)$ captures the shared visual context and co-occurrence patterns between the subject and object; (2) the subtraction term $(o_i - o_j)^2$ explicitly amplifies their spatial and appearance differences, which is a strong cue for determining spatial predicates.

Finally, the complete representations of the subject $o_i$, predicate $p$, and object $o_j$ are concatenated and fed into a classifier to predict the specific relationship category, thereby forming the final structured relational triple. This structured construction ensures that the relational semantics are deeply grounded in both general linguistic priors and fine-grained visual interactions.

\noindent\textbf{Vector Database Construction:}
To enable efficient information retrieval, we organize the identified objects, attributes, and relationships into structured text chunks. Each chunk encapsulates the semantic label, quantity, spatial location, and relational triples associated with a specific object category. To facilitate semantic matching, each text chunk is mapped into a high-dimensional dense vector using a pre-trained embedding model. These vectors $V = \{v_1, v_2, ..., v_n\}$ are then stored in a vector database, serving as the knowledge foundation for downstream reasoning.

\subsection{Semantic-Enhanced Prompt Module}
To bridge the structured visual knowledge and the LLMs, we propose a Semantic-Enhanced Prompt Module. This module is designed to integrate highly relevant visual content with user queries in a semantically clear and logically organized manner. The operation of this module consists of two main steps: knowledge retrieval and prompt formulation.

First, the textual user query $Q$ is encoded into a dense vector $\mathbf{q}$ in the same vector space as the database. The module retrieves the top-$k$ most relevant visual chunks based on cosine similarity:
\begin{equation}
\text{Top-}k = \arg\max_{v_i \in V}^{k} \left( \frac{\mathbf{q}^\top v_i}{\|\mathbf{q}\|\|v_i\|} \right),
\end{equation}

where we set $k=4$. This mechanism guarantees that the retrieved context $C = \{c_1, c_2, ..., c_k\}$ contains the visual knowledge most semantically aligned with the user's intent.

Next, the final prompt $P$ is systematically assembled. Instead of directly concatenating raw features, the prompt integrates three essential components: (1) a \textbf{prompt head} $H$, serving as a meta-prompt that explicitly instructs the LLM on how to parse the incoming structured data; (2) the \textbf{structured information} $\phi(C)$, which formats the retrieved chunks into readable text; and (3) the \textbf{user query} $Q$. The construction is formalized as:
\begin{equation}
P = H \oplus \texttt{"Context:"} \oplus \phi(C) \oplus \texttt{"Question:"} \oplus Q,
\end{equation}
where $\oplus$ denotes string concatenation. This structural design significantly improves the model's interpretability and ability to capture cross-modal relationships.

\subsection{LLM-based VQA Module}
In the final step, the meticulously constructed semantic-enhanced prompt $P$ is fed into the LLMs to generate an accurate, context-aware answer. In our implementation, we utilize Qwen-2-72B-Instruct. Importantly, our framework is highly adaptable and can be integrated with various LLMs capable of understanding structured prompts.

The answer generation process is formalized as:
\begin{equation}
A = \mathcal{L}(P),
\end{equation}
where $\mathcal{L}$ denotes the LLM's reasoning function. 

This semantic enhancement mechanism fundamentally transforms how language models process multimodal inputs by addressing linguistic ambiguity and contextual insufficiency. As the model reasons through a question, it dynamically references the curated visual knowledge $\phi(C)$. This capability is particularly crucial in dense visual environments where multiple objects interact or fine-grained details dictate the correct interpretation.

Ultimately, our approach goes beyond simple feature concatenation to achieve true cross-modal semantic alignment. By accurately grounding linguistic concepts in explicit visual relationships, the system successfully mitigates hallucination and delivers robust performance, even when faced with complex, real-world scenes involving partial occlusions and intricate spatial configurations.

%% file: Sec/4_experiments.tex
\begin{table*}[!t]
    \centering
    \setlength{\tabcolsep}{5pt}
    \caption{The Recall and F1-scores of different methods on the AUG test set. The best results are highlighted in bold. The second-best results are indicated with underlines. }
    \label{tab:AUG_results}
    \resizebox{\linewidth}{!}{
    \begin{tabular}{l cccc cccc}
        \toprule
        \multirow{2}{*}{\textbf{Method}} & \multicolumn{4}{c}{\textbf{Recall}} & \multicolumn{4}{c}{\textbf{F1-score}} \\
        \cmidrule(lr){2-5} \cmidrule(lr){6-9}
        & \textbf{Category} & \textbf{Quantity} & \textbf{Location} & \textbf{Relation} & \textbf{Category} & \textbf{Quantity} & \textbf{Location} & \textbf{Relation} \\
        \midrule
        GPT-4o~\cite{gpt4}                       & 0.1545 & 0.0947 & 0.0506 & 0.0113 & 0.2265 & 0.1365 & 0.0701 & 0.0139 \\
        Qwen-VL-Max~\cite{qwen-vl}               & \underline{0.2073} & \underline{0.1141} & \underline{0.0790} & \underline{0.0124} & \underline{0.2831} & \underline{0.1540} & \underline{0.0925} & \underline{0.0177} \\
        Gemini Flash~\cite{gemini}               & 0.1344 & 0.0814 & 0.0579 & 0.0102 & 0.2160 & 0.1272 & 0.0883 & 0.0166 \\
        LAMM~\cite{lamm}                         & 0.0609 & 0.0184 & 0.0000 & 0.0000 & 0.0819 & 0.0245 & 0.0000 & 0.0000 \\
        Internvl2.5~\cite{chen2024expanding}     & 0.1671 & 0.0968 & 0.0487 & 0.0047 & 0.2444 & 0.1403 & 0.0721 & 0.0073 \\
        QvQ-72B~\cite{qvq-72b-preview}           & 0.1331 & 0.0666 & 0.0310 & 0.0009 & 0.1853 & 0.0922 & 0.0415 & 0.0013 \\
        \midrule
        \textbf{AeroRAG (Ours)}     & \textbf{0.6494} & \textbf{0.3119} & \textbf{0.3311} & \textbf{0.1684} & \textbf{0.7156} & \textbf{0.3350} & \textbf{0.3963} & \textbf{0.1184} \\
        \bottomrule
    \end{tabular}
    }
\end{table*}

\section{Experiments}

\subsection{Experimental Setups}

\textbf{Datasets.} We evaluate our approach on three datasets collected from real-world settings. (1) AUG Dataset ~\cite{aug} comprises 400 aerial urban images captured from a top-down perspective. Characterized by densely packed, small objects and intricate relationship patterns, it is a challenging resource for research. The dataset contains 77 object categories and 63 relationship types. On average, each image includes 63 objects and 42 relationships. (2) Visual Genome (VG) dataset ~\cite{vg} comprises 108,077 images, predominantly from a first-person perspective. This viewpoint results in prominent foreground objects and significant size variations based on distance, enhancing scene depth and realism. We employ the VG-150 subset~\cite{vg150} to reduce the sparsity of rare classes. This subset contains the 150 most frequent object categories and 50 predicates, and it averages 38 objects and 22 relationships per image. (3) VQAv2 dataset~\cite{vqav2} serves as a comprehensive benchmark for evaluating general VQA capabilities. It consists of 265,016 images, with each image accompanied by at least three questions. Aligning with our framework's focus on fine-grained perception and spatial reasoning, we specifically evaluated question subsets targeting existence, quantification, and localization.

\textbf{Implementation Details.} 
We train the SGG model for 20 epochs using the Adam optimizer with a learning rate of 0.001. The training is conducted with a batch size of 64 on a machine equipped with 2 $\times$ NVIDIA RTX 3090 GPUs. For inference, the model is deployed on a server with 8 $\times$ NVIDIA RTX 3090 GPUs.

To build the vector database, we utilize the pre-trained text2vec-base model to map the structured text chunks into 768-dimensional dense vectors. During inference, the textual user query is encoded into the same vector space using this identical model. We employ the FAISS library for efficient nearest-neighbor search, retrieving the top-$k$ ($k=4$) most relevant visual chunks based on cosine similarity.

\textbf{Evaluation Metrics.} To evaluate model performance on VQA, we employ three metrics. First, recall measures the proportion of ground-truth objects, quantities, locations, and relationships correctly identified in a model's answer for an image. Second, F1-scores combine recall and precision to further assess the accuracy of these identified elements. Additionally, for the VQAv2 dataset, we adopt the standard VQA accuracy metric for performance evaluation.

\subsection{Main Results}
\begin{table*}[!t]
    \centering
    \setlength{\tabcolsep}{5pt}
    \caption{The Recall and F1-scores of different methods on the VG-150 test set.}
    \label{tab:VG150_results}
    \resizebox{\linewidth}{!}{
    \begin{tabular}{l cccc cccc}
        \toprule
        \multirow{2}{*}{\textbf{Model}} & \multicolumn{4}{c}{\textbf{Recall}} & \multicolumn{4}{c}{\textbf{F1-score}} \\
        \cmidrule(lr){2-5} \cmidrule(lr){6-9}
        & \textbf{Category} & \textbf{Quantity} & \textbf{Location} & \textbf{Relation} & \textbf{Category} & \textbf{Quantity} & \textbf{Location} & \textbf{Relation} \\
        \midrule
        GPT-4o~\cite{gpt4}                         & 0.3017 & \underline{0.1936} & 0.0734 & 0.0404 & 0.3543 & \underline{0.2198} & \underline{0.0860} & 0.0394 \\
        Qwen-VL-Max~\cite{qwen-vl}                 & \underline{0.3180} & 0.1671 & \underline{0.0793} & 0.0431 & \underline{0.3652} & 0.1922 & 0.0831 & 0.0415 \\
        Gemini Flash~\cite{gemini}                 & 0.2355 & 0.1443 & 0.0583 & 0.0252 & 0.3222 & 0.1993 & 0.0753 & 0.0325 \\
        LAMM~\cite{lamm}                           & 0.2475 & 0.1725 & 0.0000 & 0.0059 & 0.2696 & 0.1785 & 0.0000 & 0.0088 \\
        Internvl2.5~\cite{chen2024expanding}       & 0.2763 & 0.1631 & 0.0630 & \underline{0.0441} & 0.3416 & 0.1994 & 0.0774 & \underline{0.0424} \\
        QvQ-72B~\cite{qvq-72b-preview}             & 0.3030 & 0.1643 & 0.0543 & 0.0154 & 0.3299 & 0.1740 & 0.0557 & 0.0154 \\
        \midrule
        Ours (*)     & \textbf{0.5503} & \textbf{0.3468} & \textbf{0.1308} & \textbf{0.1172} & \textbf{0.4973} & \textbf{0.3056} & \textbf{0.1443} & \textbf{0.0459} \\
        \bottomrule
    \end{tabular}
    }
\end{table*}

\subsubsection{Results on the AUG Dataset}
Our main hypothesis is that a structured intermediate interface is particularly beneficial when answering depends on explicit quantities, coarse spatial grounding, and inter-object relations in dense aerial scenes. The results on AUG support this hypothesis. As shown in Table~\ref{tab:AUG_results}, AeroRAG achieves the best performance across all four attributes in both Recall and F1-score, outperforming all six baseline MLLMs on this challenging aerial benchmark.

The improvements in both metrics validate the effectiveness of our framework from two complementary perspectives. Firstly, the marked leap in Recall, such as category reaching 0.6494 compared to 0.2073 for Qwen-VL-Max, demonstrates that AeroRAG successfully mitigates the omission of small objects in dense scenes. By explicitly indexing object categories, quantities, and spatial layouts into the vector database via the Scene Graph, our retrieval mechanism helps ensure that task-critical fine-grained details are captured, contributing to a more comprehensive perception of the aerial scene.

Secondly, the clear superiority in F1-score indicates a notable enhancement in Precision. Since F1 is the harmonic mean of Precision and Recall, achieving a high F1-score, such as 0.7156 for Category, implies that the model not only retrieves more correct elements but also generates fewer hallucinated ones. This precision improvement stems from our core design: by completely replacing dense visual tokens with compact, query-relevant structured text chunks, the framework filters out background noise. This encourages the LLMs to generate answers largely grounded in the retrieved relational triples, thereby reducing visual hallucinations.

More importantly, the gains are particularly evident on attributes that are closely related to structured grounding, especially location and relation reasoning. For instance, the F1-score for Relation improves from 0.0177 to 0.1184. This trend is largely consistent with our design motivation: converting unstructured pixels into explicit relational triples $(o_i, p, o_j)$ and coordinates provides a more grounded interface for the LLMs, effectively bypassing the spatial redundancy that challenges end-to-end MLLMs in top-down aerial views.

\subsubsection{Results on the VG-150 Dataset}
To verify that the proposed framework is not limited to aerial imagery, we further evaluate it on the general-domain VG-150 dataset. As shown in Table~\ref{tab:VG150_results}, AeroRAG achieves strong overall performance among the compared methods, achieving the highest scores on all Recall metrics and all F1-scores.

Similar to the AUG results, the detailed metrics reveal that AeroRAG maintains a good balance between Recall and F1. Recall gains confirm that the scene graph interface helps preserve key visual entities in first-person images, while F1 gains indicate that the retrieved chunks provide precise grounding and help prevent overgeneration.

At the same time, the performance margin on the VG-150 dataset is notably smaller than that on the AUG dataset. For example, the Category F1-score improves by approximately 0.13 on VG-150, compared to a 0.43 improvement on AUG. This trend is in line with our design hypothesis rather than a contradiction to it. Compared with dense aerial scenes, general-domain images usually contain larger foreground objects and less cluttered layouts, where conventional dense visual tokens are often sufficient for coarse grounding. In this setting, the advantage of an explicit structured interface becomes less dramatic but remains broadly beneficial. This suggests that AeroRAG is particularly effective when the task depends heavily on quantities, spatial locations, and inter-object relations, while it is also compatible with standard visual reasoning scenarios.

\begin{table}[!t]
    \centering
    \renewcommand{\arraystretch}{1.2} 
    \caption{Performance comparison on the VQAv2 datasets.}
    \label{tab:vqav2_results}
    \setlength{\tabcolsep}{6mm} 
    \begin{tabular}{lc}
        \toprule
        \textbf{Method} & \textbf{Accuracy} \\
        \midrule
        OFA~\cite{ofa}             & 0.734 \\
        BEIT-3~\cite{beit3}        & 0.762 \\
        BLIP~\cite{blip}           & 0.821 \\
        ONE-PEACE~\cite{one-peace} & 0.825 \\
        mPLUG~\cite{mplug}         & \underline{0.857} \\
        \midrule
        \textbf{Ours}              & \textbf{0.869} \\
        \bottomrule
    \end{tabular}
\end{table}

\begin{figure}[t!]
\centering
\includegraphics[width=\linewidth]{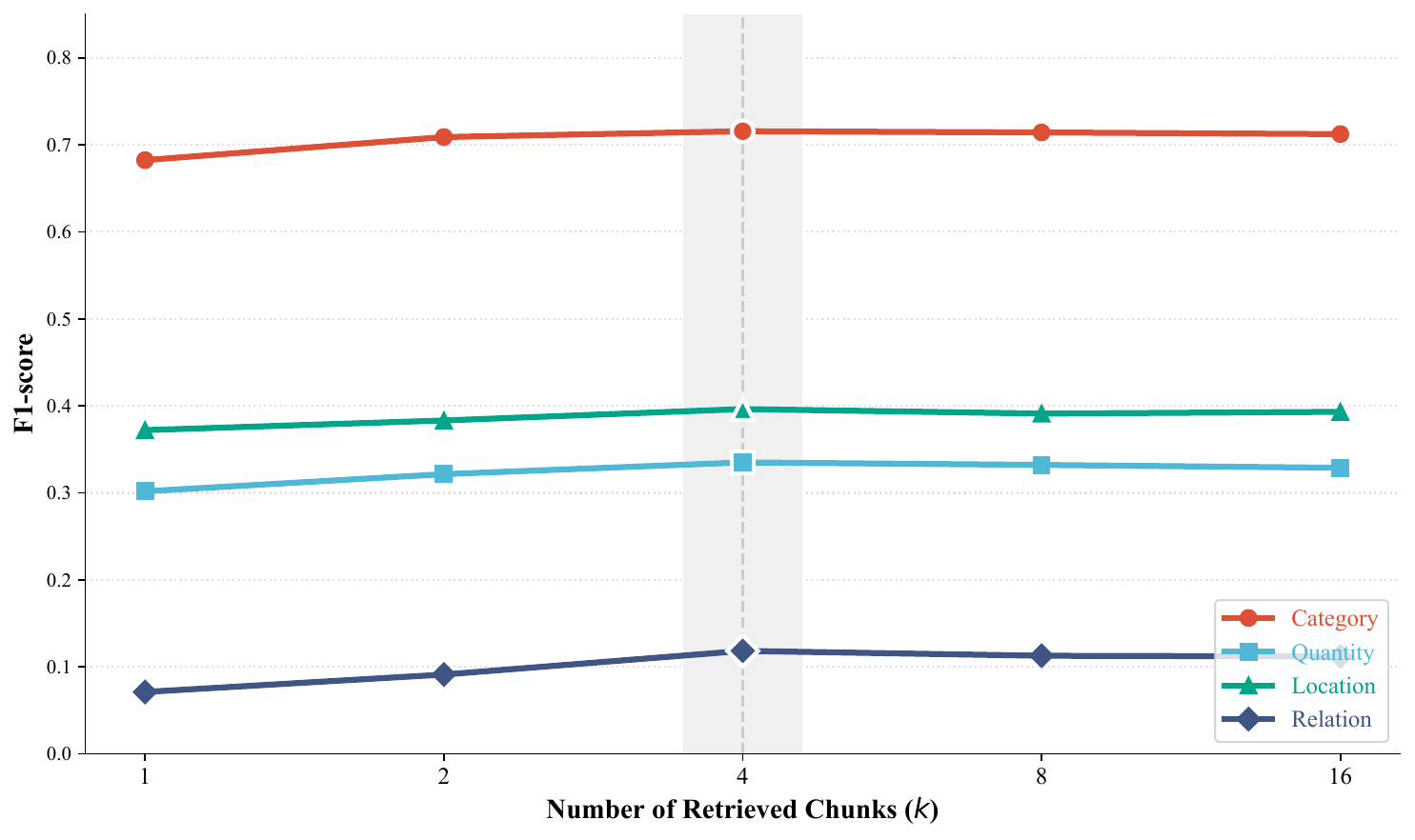}
\vspace{-2.3em}
\caption{The Ablation study on the Top-$k$ retrieval size. The F1-scores across all visual attributes peak at $k=4$, demonstrating the optimal balance between providing explicit structural grounding and avoiding visual noise in the text prompt.}
\label{fig:ablation_topk}
\vspace{-1em}
\end{figure}

\subsubsection{Results on the VQAv2 Dataset}
To further validate the generalizability of our interface beyond specialized aerospace data, we evaluate its performance on the standard VQAv2 dataset. As shown in Table~\ref{tab:vqav2_results}, our approach achieves an accuracy of 0.869, surpassing several classic VQA models.

\subsection{Ablation Studies}
\textbf{Impact of the Top-$k$ Retrieval Size.}
We investigate the influence of the retrieval size $k$ in our Semantic-Enhanced Prompt Module. 
We vary $k \in \{1, 2, 4, 8, 16\}$ and report both F1-scores and average inference latency on the AUG dataset in Table~\ref{tab:ablation_topk}, with the F1 trends visualized in Fig.~\ref{fig:ablation_topk}.

As $k$ increases from 1 to 4, all four attributes improve consistently, with the most notable gain in Relation (from 0.0712 to 0.1184). 
This confirms that retrieving more structured context progressively strengthens the explicit grounding for the LLMs, particularly for relation reasoning that heavily depends on inter-object triples.
When $k$ goes beyond 4, however, most metrics saturate or slightly decline: e.g., Quantity drops from 0.3350 to 0.3288, and Relation falls to 0.1120. 
This degradation indicates that excessive chunks introduce question-irrelevant or redundant information, distracting the LLMs and harming answer precision.

Note that the latency grows significantly with $k$, increasing from 0.82\,s for $k{=}1$ to 6.27\,s for $k{=}16$, which is attributed to the increased prompt length and the quadratic complexity of self-attention. Moving from $k{=}4$ to $k{=}8$ nearly doubles the latency, from 1.97\,s to 3.54\,s, while yielding only marginal performance gains, and $k{=}16$ is more than three times slower than $k{=}4$ with slightly inferior results. Therefore, $k{=}4$ achieves the optimal balance: it delivers the highest F1-scores across all attributes while maintaining a moderate inference cost. We adopt this setting in all subsequent experiments.

\begin{table}[t!]
    \centering
    \small
    \setlength{\tabcolsep}{4pt}
    \caption{Ablation study on the Top-$k$ retrieval size in terms of F1-scores and latency on AUG.}
    \label{tab:ablation_topk}
    \resizebox{\columnwidth}{!}{%
    \begin{tabular}{c ccccc}
        \toprule
        \textbf{Top-$k$} & \textbf{Category} & \textbf{Quantity} & \textbf{Location} & \textbf{Relation} & \textbf{Latency (s)} \\
        \midrule
        $k=1$  & 0.6824 & 0.3018 & 0.3721 & 0.0712 & 0.82 \\
        $k=2$  & 0.7089 & 0.3215 & 0.3832 & 0.0912 & 1.21 \\
        $k=4$  & \textbf{0.7156} & \textbf{0.3350} & \textbf{0.3963} & \textbf{0.1184} & 1.97 \\
        $k=8$  & 0.7142 & 0.3321 & 0.3911 & 0.1128 & 3.54 \\
        $k=16$ & 0.7123 & 0.3288 & 0.3931 & 0.1120 & 6.27 \\
        \bottomrule
    \end{tabular}
    }
\end{table}

%% file: Sec/5_conclusion.tex
\section{Conclusion and Discussion}

This paper presents AeroRAG, a scene-graph-guided multimodal retrieval-augmented generation framework for visual question answering in dense scenes. Instead of relying on direct reasoning over dense visual tokens, the proposed framework introduces a structured intermediate interface in which object categories, quantities, spatial locations, and semantic relations are explicitly represented and then selectively retrieved according to the user query. This design provides more grounded and inspectable context for downstream language reasoning, especially in aerial scenes where small objects, crowded layouts, and complex spatial dependencies make direct multimodal reasoning more difficult.

The experimental results on the AUG aerial dataset and the general-domain VG-150 benchmark demonstrate that AeroRAG consistently improves Recall and F1-score over strong multimodal baselines, with the clearest gains observed in dense aerial reasoning tasks. A supplementary result on VQAv2 further suggests that the proposed interface remains applicable beyond the aerial setting. Taken together, these findings indicate that structured retrieval is a practical way to support grounded visual reasoning across different scenarios.

More broadly, this work suggests that, for deployment-oriented visual reasoning systems, improving the interface between perception and language reasoning can be as important as scaling the multimodal backbone itself. By exposing query-relevant structured evidence to the language model, AeroRAG offers a modular and deployment-friendly alternative to direct dense-token reasoning. Despite its effectiveness, the proposed framework has certain limitations. Its overall performance inherently depends on the accuracy of upstream scene graph generation, as the results of LLMs rely entirely on the outcomes of scene graph generation, where detection or relation extraction errors may cascade into the downstream reasoning stage. Additionally, the current pipeline is restricted to static images and incurs extra computational overhead for structured knowledge construction, which poses challenges for real-time applications. In future work, we will explore extensions to dynamic scenes and more efficient retrieval backends for real-time applications.